\documentclass{article}

\usepackage{microtype}
\usepackage{graphicx}
\usepackage{subfigure}
\usepackage{booktabs} 

\usepackage{amsmath}
\usepackage{amssymb}

\newcommand{\R}{\mathbb{R}}
\usepackage{graphics}
\usepackage{multirow}
\usepackage{float}
\usepackage{relsize}
\usepackage{enumitem}
\usepackage{booktabs}
\usepackage{pdfpages}
\usepackage{algorithm}
\usepackage{algorithmic}
\usepackage{varwidth}
\newcommand{\RNum}[1]{\uppercase\expandafter{\romannumeral #1\relax}}
\usepackage{enumitem}
\usepackage{hyperref}

\usepackage[accepted]{mlsys2021}

\mlsystitlerunning{Don't Forget to Sign the Gradients!}

\begin{document}

\twocolumn[
\mlsystitle{Don't Forget to Sign the Gradients!}
\begin{mlsysauthorlist}
\mlsysauthor{Omid Aramoon}{umd}
\mlsysauthor{Pin-Yu Chen}{ibm}
\mlsysauthor{Gang Qu}{umd}
\end{mlsysauthorlist}

\mlsysaffiliation{umd}{Department of Electrical and Computer Engineering and Institute of Systems Research, University of Maryland}
\mlsysaffiliation{ibm}{IBM Research}

\mlsyscorrespondingauthor{Omid Aramoon}{oaramoon@umd.edu}

\mlsyskeywords{Machine Learning, MLSys}

\vskip 0.3in

\begin{abstract}
Engineering a top-notch deep learning model is an expensive procedure that involves collecting data, hiring human resources with expertise in machine learning, and providing high computational resources. For that reason, deep learning models are considered as valuable \textbf{Intellectual Properties} (IPs) of the model vendors. To ensure reliable commercialization of deep learning models, it is crucial to develop techniques to protect model vendors against IP infringements. One of such techniques that recently has shown great promise is \textbf{digital watermarking}. However, current watermarking approaches can embed very limited amount of information and are vulnerable against watermark removal attacks. In this paper, we present GradSigns, a novel watermarking framework for deep neural networks (DNNs). GradSigns embeds the owner's signature into the gradient of the cross-entropy cost function with respect to inputs to the model. Our approach has a negligible impact on the performance of the protected model and it allows model vendors to remotely verify the watermark through prediction APIs. We evaluate GradSigns on DNNs trained for different image classification tasks using CIFAR-10, SVHN, and YTF datasets. Experimental results show that GradSigns is robust against all known counter-watermark attacks and can embed a large amount of information into DNNs. 
\end{abstract}
]

\printAffiliationsAndNotice{\mlsysEqualContribution} 

\section{Introduction}
Designing a deep neural network (DNN) involves the costly process of training numerous models with different architectures and hyperparameters to find an optimal model with superior performance. Depending on the complexity of these models and their targeted task, each of these rounds of trial-and-error training can take several days to weeks to complete on cutting-edge GPUs. Moreover, gathering large enough training data can be an expensive task, which only adds up to the design cost of such models. Therefore, 
top-level DNNs are considered as the intellectual property of model vendors.

Extensive applications of DNNs in different sectors have opened a new market for these products. Companies are either commercializing their models similar to software and mobile Apps, or are monetizing the prediction capabilities of their products on \emph{Machine learning as a Service (MLaaS)} platforms \cite{ribeiro2015mlaas}. In such a setting, adversaries can purchase licensed models and illegally redistribute them over the market for lower prices, and take the market share from legitimate vendors. Moreover, licensed customers might deploy purchased models in their future products without asking for permission and paying the copyright fees, which would diminish the profit margin of model vendors. Therefore, there is an urgent need for finding ways to protect vendors against IP infringements. It is clear that reliable commercialization of deep learning models is not possible unless vendors are able to verify and prove ownership of their products. 

\begin{table*}[t]
  \caption{Properties of an effective watermarking technique for deep learning models. \newline}
  \label{requirments}
  \centering
  \resizebox{0.8\textwidth}{!}{
  \begin{tabular}{|l|l|}
    \hline
    Properties & Description \\
    \hline
     Loyalty & Watermark should have negligible overhead on the model's performance.\\
     \hline
     Robustness & Watermark must remain verifiable in the presence of anti-watermark attacks.\\
     \hline
     Reliability & Watermark verification should result in minimal false ownership claims.\\
     \hline
     \multirow{1}{*}{Credibility}&  Finding or forging a fake (ghost) watermark should not be feasible. \\
     \hline
     Efficiency & Watermark extraction and verification should incur low costs.\\
     \hline
     Capacity & Watermarking technique should be able to embed large signatures.\\
    \hline
  \end{tabular}
  }
\end{table*}

Digital watermarking has been extensively used for proving ownership and preventing piracy of a variety of IPs such as multimedia and video contents \cite{borra2018digital,saini2014survey,visionWM}, computer software \cite{softwarewmsurvey}, Integrated Circuits (IC) \cite{Qu_book03,Wong04}, etc. Proving ownership of an IP with digital watermarking is a two-step process: \emph{embedding} and \emph{verification}. In the embedding phase, the IP owner uses a watermark embedding algorithm to insert their signature into their IP, which in our case, is the deep learning model. The watermark is successfully embedded if the owner can fully retrieve their embedded watermark using the extraction algorithm. If the owner suspects that some model might belong to them, they can investigate and prove the ownership by extracting the potential watermark in the suspicious model, and comparing it with their own signature. Table \ref{requirments} lists the properties that an effective watermarking technique for deep learning models should have \cite{rouhani2018deepsigns,Guo,adi2018turning}. 

Recently, digital watermarking of DNNs has gained attention in the research community, and several watermarking methods \cite{Uchida_2017,namba2019robust,rouhani2018deepsigns,merrer,Zhang,Guo,adi2018turning} have been proposed. The existing watermarking methods are categorized into white-box and black-box, according to their assumption about the setting of the verification phase. White-box watermarking methods assume that the model vendor has full access to the internal details (such as architecture, hyper-parameters and weights) of the supposedly stolen model to investigate and prove their ownership. Black-box watermarking methods, on the other hand, assume that the stolen model will only be accessible over APIs or as an encrypted software; therefore, the model vendor has only access to model's output to investigate the ownership. Black-box methods are more desired as adversaries are unwilling to provide the public with white-box access to the stolen models in the fear of getting caught by law enforcement.

In this paper, inspired by the constraint-addition based watermarking for IC designs \cite{Qu98, Qu_book03}, we introduce GradSigns, a novel digital watermarking framework for DNNs. GradSigns embeds the watermark information by imposing a statistical bias on the expected gradients of the cost function with respect to the model's input. Notably, GradSigns has little or no impact on the prediction accuracy of the marked model, and can be verified over API access. More importantly, unlike contemporary black-box watermarking methods, GradSigns is able to embed more than one bit of watermark information into DNNs, and is also shown to be extremely robust against \textit{general} counter-watermark methods (such as parameter pruning, model fine-tuning and query invalidation), and possible \textit{adaptive} watermark removal techniques which are designed with complete knowledge of our framework.

Our main \textbf{contributions and findings} are as follows,
\begin{itemize}[topsep=0pt,leftmargin=*]
    \item[$\checkmark$] We propose GradSigns, a novel and robust watermarking framework for DNNs. To the best of our knowledge, GradSigns is the first watermarking method leveraging gradient of model's components to embed owner's signature.
    \item[$\checkmark$] We evaluate the performance of GradSigns on DNNs trained for three different image classification tasks including object and hand-written digit classification, and face recognition. Our evaluations reveal that GradSigns is able to successfully embed watermark information with negligible impact on the model's performance. We also show that GradSigns is highly effective and reliable in protecting the owner's copyright.
    \item[$\checkmark$] We demonstrate thorough rigorous experiments that unlike contemporary black-box watermarking methods, our framework can embed a large amount of information into DNNs, and watermarks embedded by GradSigns are robust against known and proposed adaptive counter watermark attacks mounted by strong adversaries.
\end{itemize}

\section{Related Work and Threat Model}

\subsection{Related Work}
The study by \citet{Uchida_2017} was the first work bringing attention to watermarking of DNNs. They proposed a white-box watermarking method that embedded the watermark into the parameters of layers in the model.

\citet{rouhani2018deepsigns} presented DeepSigns, a watermarking framework that can be applied to both white-box and black-box settings. In the black-box version of DeepSigns, the unmarked model gets fine-tuned using a mixture of original training data and a set of crafted images with random labels. Among the images in the set, the ones that are misclassified by the unmarked model and correctly classified by the fine-tuned model (the watermarked model), are selected as the \emph{watermark key set}. The model belongs to the owner, if it shows high accuracy on her watermark key set.

Note that the other existing black-box watermarking methods follow a similar procedure to embed the watermark into the host model. The host model is trained to behave in a pre-defined and rather unusual way upon encountering input samples from a set of specially crafted images, a.k.a the watermark key set. This unusual behavior differentiates marked models from similar unmarked models, and allows model vendors to identify their designs.
 
\citet{merrer} introduced a black-box watermarking technique that uses true and false adversarial examples as the watermark key set. A true adversarial example is a sample superimposed with small and intentional perturbations that causes a model to misclassify the sample. On the other hand, a false adversarial example is a sample superimposed with adversarial perturbation, and yet classified correctly by the model. 

\citet{Zhang} proposed a black-box watermarking framework that uses especially crafted samples as the watermark key set. They used three different methods to generate the watermark key set: (1) selecting unrelated images from another dataset, (2) superimposing images with a meaningful content (sticker, text, etc.) or (3) superimposing images with a Gaussian noise. In their method, watermark key samples are assigned random labels, and are used alongside the original training set to train the model. \citet{adi2018turning} proposed a similar watermarking technique in which abstract images were used as the watermark key set. Another similar approach can be found in \citet{Guo}.

\citet{namba2019robust} proposed a black-box watermarking method specifically designed to be robust to parameter pruning. They argued that to ensure the resiliency of the watermark against parameter pruning attack, the watermark should be embedded via parameters of the model that have large absolute values and therefore, significantly contribute to the original classification task. To embed the watermark, they leveraged a custom activation function that first exponentially weights incoming parameters of each layer, and then calculates the sum of weighted inputs for each neuron. They used a set of random images with random labels as their watermark key set.
\subsection{Threat Model}
\label{threatmodel}
The threat model includes two parties \emph{\textbf{model vendor}} and \emph{\textbf{adversary}}. The model vendor owns model $M$, a DNN which they have engineered and trained for a certain task $T$ using the dataset $D$. Dataset $D$ is collected and owned by the model vendor. The second party, the adversary, is an entity that doesn't have the required resources for designing and training a top-notch model, and wishes to make a profit out of model $M$ without paying any copyright fee to the model vendor. The adversary can be a company that has purchased the license of $M$ for one of their products and want to deploy it on another one without paying additional copyright fees. They can also be any entity who somehow has got their hands on the model, and wish to sell it on the darknet. 
Model vendor's goal is to protect $M$ against IP infringements by 
means that enable the vendors to prove their ownership, and possibly detect the source of theft. On the other hand, the adversary's ultimate goal is to continue profiting from $M$ without getting caught by law enforcement. 

In our threat model, we assume the strongest adversary who has the expertise and the computation power required for training a model. However, the dataset that they have available for task $T$ is far smaller than dataset $D$ owned by the model vendor, and therefore is not large enough for them to train a top-grade model from scratch. If the adversary had access to dataset $D$, they would not need to hijack $M$ as they are capable of training the model themselves. Similar to prior arts, we assume that the adversary is capable of trying any of the general anti-watermark attacks such as parameter pruning, model fine-tuning, and query invalidation and modification to remove the watermark or obstruct verification. In addition, we assume that the adversary is aware of all existing watermarking techniques, and is capable of designing adaptive anti-watermark schemes to target the deployed method. Due to such possible counter watermark attempts, it is safe to assume that a hijacked model will go under modifications before being monetized by the adversary. The adversary accomplishes their goal, if they can remove the vendor's signature or obstruct watermark verification without sacrificing too much on the performance of the model. Note that a counter watermark attempt that drastically degrades the model's performance is not considered successful. 

\begin{figure*}
    \centering
    \includegraphics[width=\linewidth]{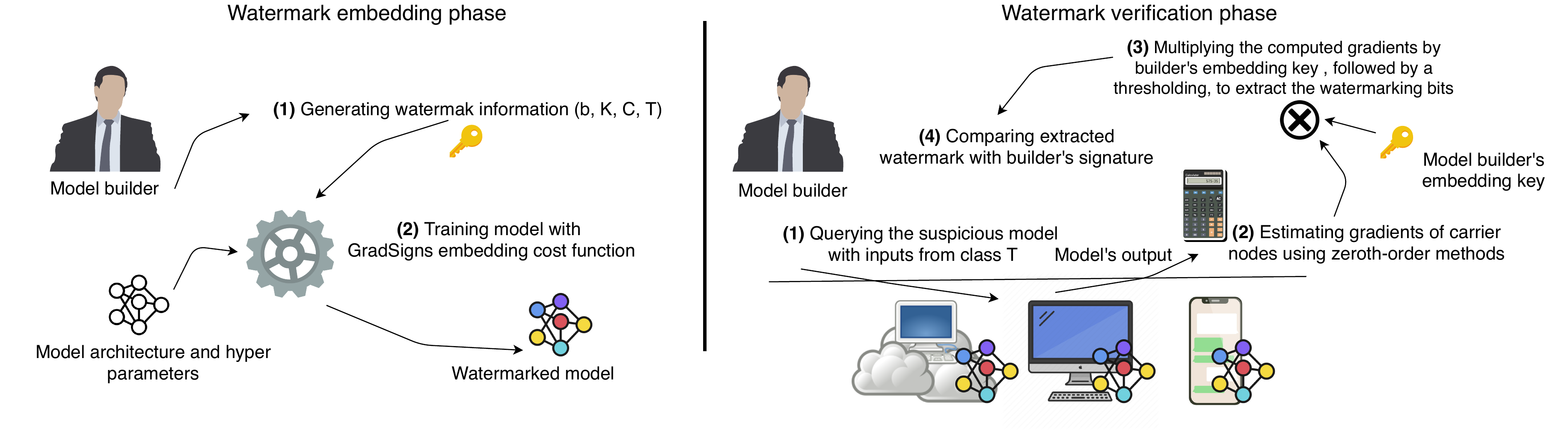}
    \caption{Workflow of watermark embedding and verification using GradSigns.}
    \label{workflow}
\end{figure*}

\section{GradSigns Watermarking Framework}
\label{sec_GradSigns}
The foundation of our proposed watermarking framework, GradSigns, is based on the fact that there can be more than one unique solution to the non-convex optimization problems that deep learning models are designed to solve. Deep learning models, mainly due to their over-parametrization, transform the loss surface to have a manifold of optimal solutions which enables optimization algorithms to find multiple solutions yielding comparable performances on the testing data set \cite{du2018power,allen2019learning}. For classification tasks, each solution corresponds to a unique decision boundary with similar classification accuracy.

Intuitively, GradSigns finds a solution, i.e. a set of model parameters, corresponding to a decision boundary that not only results in comparable performance on the original task but also fulfills an additional goal which is carrying the owner's watermark information. GradSigns works by embedding watermark information into the expected gradient of the cross-entropy cost function with respect to the model's input. For any input sample $x$, the gradient of the cost function with respect to the input is a vector tangent to the model's cost function surface, and perpendicular to the decision boundary at point $x$. Therefore, by imposing a statistical bias on these gradients, GradSigns is essentially reshaping the decision boundary to incorporate the desired watermark information. For simplicity, we refer to the gradient of the cross-entropy cost function with respect to the input of the model as \textit{gradient of input} or \textit{input gradient} in the rest of the paper.
Figure \ref{workflow} illustrates the workflow of GradSigns, and \textsection \ref{wm-embedding-section} and \ref{wm-extraction-section} describe the watermark embedding and verification phase of our watermarking framework.

\subsection{Watermark Embedding}
\label{wm-embedding-section}
Forcing a random statistical bias on the gradient of inputs to the model can drastically degrade its performance on the classification task. To ensure a successful marking without sacrificing the model's performance, the watermark needs to be embedded while optimizing the model for the original task. For that reason, similar to prior arts such as \cite{rouhani2018deepsigns,Uchida_2017}, we embed the watermark into the host model by including a regularizer term in the model's training cost function. The final training cost function including the regularizer term is defined as:
\begin{equation}
\label{GS-loss}
\resizebox{\columnwidth}{!}{
 $J(x |\,\theta,\,y) = J_{cross-entropy}(x|\,\theta,\,y) +  \lambda \,J_{Embedding}(x|\,\theta,\,y)$  }
\end{equation}
Here, $\theta$ denotes the model's parameters, $x$ is an input sample, $y$ is the ground truth label for input sample $x$, $J_{cross-entropy}(.)$ is the task-specific cost function which is the cross-entropy function for classification problems, $\lambda$ is the trade-off hyperparameter, and $J_{Embedding}(.)$ is the watermark embedding regularizer term which penalizes the distance between the expected value of input gradients and the desired watermark.
Before defining the embedding regularizer term, we explain the steps that the model vendor needs to take prior to embedding the watermark. These steps are as follows:\newline
\textbf{Step 1.} Generating an $N$-bit vector $b \in \{0, 1\}^{N}$ to be used as the watermark. \newline
\textbf{Step 2.} Randomly selecting a set $C$ of input neurons to carry the watermark. We refer to set $C$ as the \textbf{\emph{watermark carrier set}}, and to neurons in $C$ as \textbf{\emph{carrier nodes}}. The gradient of inputs observed on neurons in the career set participate in embedding the watermark. \newline
\textbf{Step 3.} Generating an \textbf{\emph{embedding key}} $K^{N \times |C|} \in [-1,1]^{N \times |C|}$. Embedding key is a transformation matrix that maps the expected gradient of carrier nodes to a binary vector of size $N$. \newline
\textbf{Step 4.} Selecting a random target class \textbf{$T$}. Images from class $T$ are used to calculate the gradients of carrier nodes.

Note that generating the watermark $b$ and embedding key $K$ can either be done randomly by using a Random Number Generator (RNG) or by hashing a message containing information that can be used to prove vendor's ownership as further explained in \textsection \ref{credibility-Sec}. 
In our method, the watermark is successfully embedded if the following property holds:
\begin{equation}
\label{embedding-success-eq}
\forall j \in \{0,1,...,N-1\},\;\;\mathlarger{\chi_{[0,\infty)}}\left(\sum_{i=0}^{|C|-1} K_{ji} G_{i}\right) = b_{j}
\end{equation}
Here, $G \in \R^{|C|}$ is the expected gradient of cross-entropy function with respect to carrier nodes in $C$, measured over a sample of images from target class $T$, $K$ is the model vendor's embedding key, $b_{j}$ is $j_{th}$ watermark bit, and $\mathlarger{\chi}$ is a step function outputting one for values greater than zero.

For each watermark bit $j$, Equation \ref{embedding-success-eq} denotes a linear inequality where the expected gradients of carrier nodes ($G$) are the variables, and the $j$-th row of the embedding key ($K$) are the coefficients, as shown in Equation \ref{embedding-equation}. This linear inequality is essentially denoting a half-space where acceptable values of expected gradients can reside for a successful embedding of watermark bit $j$.
\vspace{-0.12cm}
\begin{equation}
\label{embedding-equation}
(-1)^{b_{j}}\sum_{i=0}^{|C|-1} K_{ji} G_{i} < 0
\end{equation}
The task of embedding each watermark bit $j$ can also be viewed as a binary classification task with a single layer perceptron (SLP) where the parameters of the perceptron layer are fixed to a constant value equal to $j_{th}$ row of the embedding key ($K$), and only the input of the SLP, i.e. the gradient of the carrier nodes, are being trained. To this end, we utilize a binary cross-entropy loss function in the embedding regularizer to embed each watermark bit:
\begin{equation} 
J_{Embedding}(\theta) = -\sum_{j=0}^{N-1}(b_{j}\,log(y_{j}) + (1-b_{j})\,log(1-y_{j}))
\end{equation}
Here, $y_{j} = \sigma(\sum_{i} K_{ji} G_{i})$ is the output of the SLP corresponding to the $j_{th}$ watermarking bit, and $\sigma$ is the sigmoid function.
\vspace{0.25cm}
\subsection{Watermark Extraction}
\label{wm-extraction-section}
To extract a watermark embedded by GradSigns, the first step that the model owner needs to take is computing the expected gradient of the carrier nodes. In the white-box setting where the owner has access to the internal configurations of the suspicious model, the gradients can be calculated by backpropagation. However, in the black-box setting, computing gradients via backpropagation is not possible.

To enable watermark extraction in the black-box setting, we propose using a zeroth-order gradient estimation method to calculate the expected gradients of the carrier nodes. Zeroth order methods can estimate gradient with respect to any direction $v$ by evaluating the cost function value at two very close points located along this direction \cite{ghadimi2013stochastic,liu2020primer}.
We use the difference quotient to estimate the gradient of cost function with respect to carrier nodes as shown below:
\begin{equation}
\label{zeroth-order-grad-est-Equation}
\hat{G_{c}}(x)\,=\,\frac{\partial\,J(x)}{\partial x_{c}} \approx  \frac{J(x\,+\,he_{c})\,-\,J(x\,)}{h}
\end{equation}
where $\hat{G_{c}}(x)$ is the estimated gradient of carrier node $c$ at point $x$, $h$ is the estimation step length which is set to $0.0001$ throughout our experiments, $e_{c}$ is a standard basis vector with 1 at the component corresponding to career node $c$, and 0s elsewhere. Please note that the value of cross entropy function $J(x)$ can be computed in the black-box setting, given model's output and the ground truth label for input $x$.

In Equation \ref{zeroth-order-grad-est-Equation}, for each input $x$, the gradient of a carrier node is calculated by evaluating the value of the cost function for two points whose coordinates are the same except for the one coordinate corresponding to the carrier node. The gradient estimation error of career nodes, not including the error introduced by limited numerical precision, is in the order of $O(|C| \cdot h^{2})$ \cite{liu2018zeroth}. For any input $x$, we need to evaluate the cost function $|C|+1$ times to estimate the gradients of all carrier nodes. We note that our watermark extraction naturally applies to more query-efficient gradient estimation methods such as \cite{liu2018signsgd}.

After calculating the expected gradients for all carrier nodes, the model vendor can retrieve the embedded watermark by multiplying expected gradients with their embedding key. The model belongs to the vendor, if the Bit Error Rate (BER) of the extracted watermark is lower than a certain threshold. In \textsection \ref{Reliable-Sec}, we explain how the BER threshold is determined to assure a reliable watermark verification, i.e. low false positive and high detection rates.

\section{Evaluation}
\label{Evaluation-Section}

\begin{figure*}[t]
    \centering
    \includegraphics[height=4.5cm]{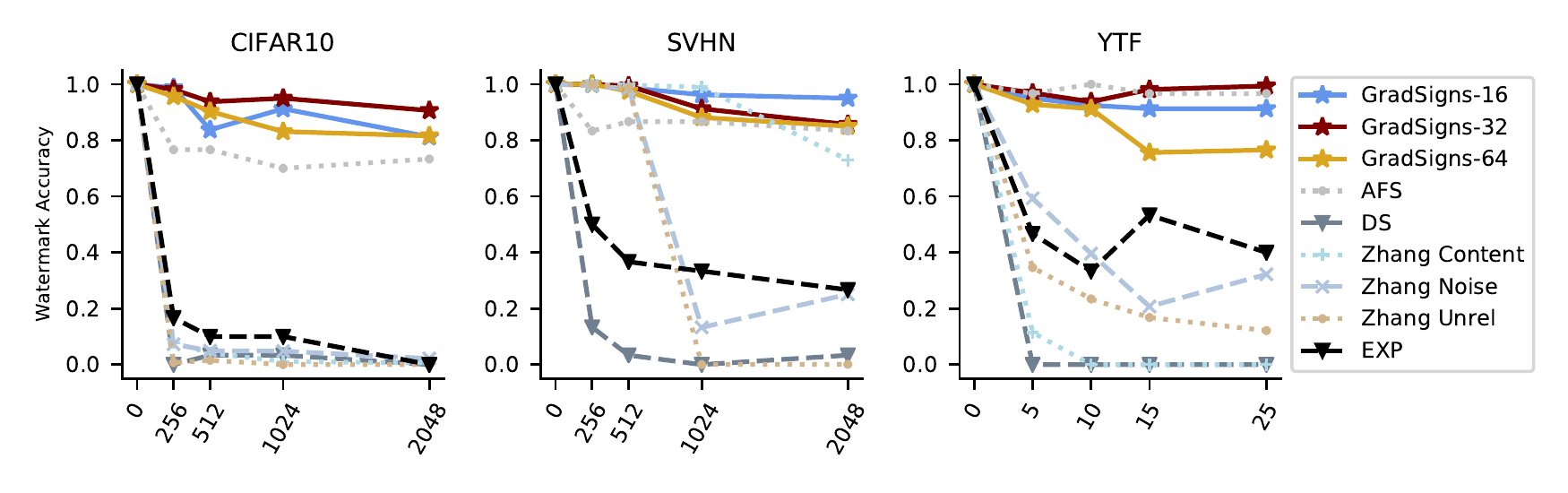}
    \caption{Watermark accuracy of contemporary watermarking methods in presence of model pruning and fine-tuning removal attacks. The horizontal axis of each diagram demonstrates the number of samples available for each classification label in adversary's dataset. }
    \label{pr-ft-diagram}
\end{figure*}

In this section, we evaluate performance of GradSigns with respect to the requirements mentioned in Table \ref{requirments} on benchmark models trained on various image datasets including CIFAR-10 \cite{cifar10}, SVHN \cite{SVHN}, and YTF \cite{wolf2011face}. For a detailed description of these datasets, please refer to Appendix \textsection \ref{dataset_app}. ResNet20 \cite{he2016deep} and Deep-ID \cite{sun2014deep} are benchmark neural networks considered for CIFAR10 and YTF datasets, respectively. For SVHN dataset, a convolutional neural network with 6 convolution layers and 2 fully connected layers is selected as the benchmark model (please refer to Appendix \textsection \ref{dataset_app} for a detailed description of this model.)      

We embed watermarks with three different sizes of 16, 32 and 64 bits using GradSigns for all benchmark models. Images of airplanes, number "1", and individual number 1 are used as the target class for CIFAR-10, SVHN and YTF benchmarks models, respectively. Trade-off hyperparameter $\lambda$ is empirically set to a value in range $(0.0,1.0]$ depending on the benchmark, size of the watermark carrier set $|c|$ and watermark length. Selecting a large $\lambda$ can degrade watermarked model's performance while choosing a small $\lambda$ leads to failure in successfully embedding the watermark. In our experiments, $\lambda$ is set to the smallest value (over trials with varying values for $\lambda$) that can successfully embed the watermark.
Moreover, our experiment showed that using larger carrier sets increases the Bit Embedding Success Rate (BESR) of embedding attempts. BESR denotes the ratio of watermark bits that are inserted successfully into the model. However, we note that a large carrier set would translate to higher watermark verification costs in the black-box setting, which will be further discussed in \textsection \ref{Efficinecy-Sec}. To this end, in our experiments, we select the smallest watermark carrier set size (over trials with varying carrier set sizes) which is capable of embedding the watermark successfully. Table \ref{efficinecy-table} reports the chosen watermark carrier set sizes for all benchmarks. 

In what follows, we examine the properties of an effective watermarking method as summarized in Table \ref{requirments}.

\subsection{Loyalty}
\label{loyalty-Section}
An ideal watermarking method should have minimal impact on the performance of the host model. Comparing the prediction accuracy of the baseline (unmarked) and marked models in Table \ref{benchmarks-test-acc-table} demonstrates that GradSigns meets the loyalty requirement as it incurs on average less than 1\% drop in the performance of all benchmark models.

SVHN and YTF baseline benchmarks are trained using the same training setting (epochs, learning rate, hyperparameters, etc.) adapted for their watermarked counterparts, and their best performance on validation sets is reported on Table \ref{benchmarks-test-acc-table}. For the CIFAR10 baseline model, we are reporting the maximum validation accuracy claimed in the original paper \cite{he2016deep} for a ResNet20 trained on the CIFAR10 dataset. GradSigns is compliant with this requirement for (1) it embeds the watermark information by simultaneously optimizing for both model's prediction accuracy and the watermark embedding regularizer mentioned in \textsection \ref{wm-embedding-section}, and (2) it only affects parts of decision boundary that pertain to samples of the target class.

\vspace{-1em}
\begin{table}[h]
  \caption{Test accuracy of watermarked models. GradSigns has negligible impact on performance of marked models. \newline}
  \label{benchmarks-test-acc-table}
  \centering
  \resizebox{0.9\columnwidth}{!}{
    \begin{tabular}{|c|c|c|c|c|c|}
    \hline
    Dataset & Baseline Acc.  &\textbf{GS-16} & \textbf{GS-32} & \textbf{GS-64}  \\
    \hline
    CIFAR-10  & 91.2\% & 90.1\% & 90.4\% & 90.5\%  \\
     \hline
     SVHN & 96.1\% & 95.5\% & 95.7\% & 95.3\%  \\
     \hline
     YTF  & 99.6\%  & 99.6\% & 99.6\% & 98.6\%  \\
     \hline
    \end{tabular}
    }
\end{table}
\vspace{-1em}

\subsection{Reliability}
\label{Reliable-Sec}
If a marked model remains untouched, the watermark extracted from the model would completely match with the one embedded by the owner. However, it is likely the marked model, before being monetized by the adversary, goes under modifications due to counter watermark attempts such as fine-tuning, parameter pruning, etc. Therefore, there is a possibility that the watermark extracted from the modified model might not completely match with the owner's signature. On the other hand, the watermark extracted from similar unmarked models might partly match with owner's signature. Therefore, to reliably determine the ownership of a model, it is crucial to establish a threshold for tolerated mismatched bits on the extracted watermark. Based on this threshold, the model owner can decide and prove whether the modified model originally belonged to them or not.

We rely on null hypothesis testing for verifying the ownership of a model. As mentioned in \textsection \ref{wm-embedding-section}, possible values of expected gradients for a successful embedding of a \emph{single} watermark bit form a half-space. Therefore, the probability that expected gradients of a null-model, any unmarked model trained for the same task, coincidentally reside in this half-space is $\frac{1}{2}$. Let $\eta$ be the maximum number of bit errors tolerated on the extracted watermark to safely claim the model's ownership. Assuming that model $M$ is a null-model, the probability that the extracted watermark from $M$ has at most $\eta$ erroneous bits is as follows:
\vspace{-0.5em}
\begin{equation}
\label{threshold-eq}
    P(n_{error} \leq \eta | M) = \sum_{k=0}^{\eta} \binom{N}{k}(\frac{1}{2})^{N-k}(1-\frac{1}{2})^{k}
\end{equation}
Here, $n_{error}$ is a random variable denoting the number of erroneous bits, and $N$ is the length of the embedded watermark. To reliably reject that a model is behaving like a null-model, we require the probability in Equation \ref{threshold-eq} to be upper bounded by $\tau$, a close to zero threshold. Solving $P(n_{error} \leq \eta | M) < \tau$ for $\eta$ yields the maximum number of bit errors tolerated on the extracted watermark to safely verify model's ownership. In all our experiments, we set the threshold $\tau$ to $3\times10^{-3}$. With $p-value\,<\,3\times10^{-3}$, minimum number of correct (matching) bits required to reliably claim ownership of a model marked with signatures of size 16, 32, 64 bits are 14, 25 and 44 bits, respectively. A watermark is declared \textbf{existent and verifiable} in a model \textit{if and only if} the extracted signature has less erroneous bits than the computed threshold $\eta$.    
\subsection{Robustness}
\label{Robustness-Section}
In this section, first, we evaluate robustness of our framework against \textit{general} counter watermark schemes such as \textit{model fine-tuning}, \textit{model pruning}, and \textit{query invalidation-modification} attacks. These anti-watermark schemes are applicable to any watermarking framework, and have been widely considered in prior arts. We also perform a thorough comparison of resiliency of GradSigns and other contemporary black-box digital watermarking methods including DS \cite{rouhani2018deepsigns}, EXP \cite{namba2019robust}, AFS \cite{merrer}, and Zhang's \cite{Zhang} against such counter watermark attacks. For a fair evaluation, we don't experiment with white-box techniques such as Uchida's \cite{Uchida_2017} or white-box DeepSigns \cite{rouhani2018deepsigns} as they benefit from stronger underlying assumptions.

Assuming that the adversary is aware of our framework, they might attempt to tamper with the gradient of carrier nodes to obstruct verification. To this end, we consider several adaptive counter watermark attacks to further solidify the evaluation of GradSign's robustness.

\subsubsection{Parameter pruning and model fine-tuning}
\label{param-prun-finetune-section}
Parameter pruning and model fine-tuning are two of the most common attacks against watermarks in literature. Fine-tuning is retraining the model's parameters with the training or a new dataset to find other local minima with better performance on the original task. Parameter pruning is a compression technique used to reduce the model's computational complexity and memory usage. It is often applied to DNN models deployed on embedded systems and mobile devices. Similar to prior arts, we utilize the parameter pruning technique in \cite{han2015learning} where $p\%$ of model parameters with smallest absolute values are set to zero and removed from the computation graph of the model. $p$ is the \emph{pruning rate} which decides the portion of model parameters to be removed. Parameter pruning is usually followed by model fine-tuning so the model can recover from the possible performance drop due to model compression. 

Figure \ref{pr-ft-diagram} shows the watermark accuracy of all seven watermarking methods in presence of model pruning and fine-tuning attacks across all three benchmarks.
Experimental setup and implementation details of these techniques are described in Appendix \textsection \ref{dataset_app}. The methods proposed in \cite{Guo,adi2018turning} are not evaluated in \textsection \ref{Evaluation-Section} as they essentially follow a similar approach as \cite{Zhang} to embed the watermark, and their contribution is not introducing a new watermarking technique but establishing a reliable watermark verification using cryptography. 

The horizontal axis of diagrams in Figure \ref{pr-ft-diagram} \emph{denote the number of samples available per classification label in adversary's dataset.} For each benchmark, we have considered adversaries with four different sizes of sub-datasets in hand. Stronger adversaries have access to datasets of comparable size to the original training dataset while weaker ones have only a small number of samples to work with. We note that an adversary with access to datasets larger than what's considered in Figure \ref{pr-ft-diagram} has no incentive to hijack the model as they are capable of training the model from scratch with comparable performances. As shown in Figure \ref{pr-ft-diagram}, \textbf{\textit{only}} watermarks embedded by GradSigns show \textbf{consistent} resiliency against "pruning+fine-tuning" attack, and remain existent and verifiable across all three benchmarks.    

In our experiments, $70\%$ of adversary's datasets is dedicated as the training set, and the rest as the validation set. Every watermarked model is pruned with 10 different pruning rates varying from 0\% to 90\% resulting in 10 different compressed models. Then, each pruned model is fine-tuned for 10 epochs with early stopping on the accuracy of validation. The learning rate is set 0.0005 which is equal to or greater than the learning rate in the final stage of training of benchmarks. The watermark accuracy reported in Figure \ref{pr-ft-diagram} is the minimum watermark accuracy observed across all the "pruned+fine-tuned" models whose test accuracy doesn't fall beneath 10\% of the baseline model's. As mentioned in \textsection \ref{threatmodel}, a removal attempt is not successful if it leads to a significant drop in the performance of the model.

\subsubsection{Query invalidation-modification}
Query invalidation and modification \cite{namba2019robust} is a counter watermark attack in which the adversary utilizes an autoencoder to identify and modify model queries that are executed to extract the watermark. While this counter watermark attack has proven effective \cite{namba2019robust} against watermarking techniques proposed by \citet{Zhang} and \citet{rouhani2018deepsigns}, it doesn't apply to GradSigns. The watermark key set for GradSigns is constructed from samples of training data without any modification or relabeling, which renders this attack futile against our method. Please refer to the Appendix \textsection \ref{qim-app} for a more detailed description of this attack.

\subsubsection{Adaptive Attacks}
Assuming that the adversary is aware of our framework, they might attempt to corrupt the estimated gradient for carrier nodes to prevent watermark verification. To achieve this goal, the adversary can take any of the following approaches: (a) \textit{Model tampering}: modifying the model parameters to change values of gradients (b) \textit{Input tampering}: modifying inputs to the model to incur error on the estimated gradients, and (c) \textit{Output tampering}: tampering reported prediction scores to corrupt approximated gradients.

Model fine-tuning and model pruning are known instances of model tampering attacks that have been thoroughly investigated in \textsection \ref{param-prun-finetune-section}. Besides these counter watermark attacks, we consider \textit{adversarial fine-tuning} and \textit{model quantization} in our experiments which fall within the same category of model tampering attack.
To further broaden the scope of our evaluations, we propose three adaptive counter watermark attacks against GradSigns, namely \textit{input tampering}, \textit{score rounding}, and \textit{score perturbation}, which are instances of input and output tampering attacks. 

For the rest of this section, we only report experimental results for 64-bits marked benchmarks due to space limitations. We note that results for other benchmarks are consistent with the statements we make in this section.   

\textbf{Model Quantization} is a compression technique to decrease model's memory footprint by reducing the number of bits that are required to represent its parameters. Along with weight pruning, model quantization is among the popular techniques for making inference more efficient in resource-limited settings. We conjectured that model quantization might tamper with the information carried over the gradients by modifying model weights, however, our experiments showed the contrary. In this experiment, we first quantized parameters of marked benchmarks to 8 fixed point integers, and then performed watermark extraction. We were able to verify the watermark successfully from all benchmarks, which showed model quantization can not remove watermarks embedded by GradSigns. Table \ref{quant-table} reports the performance overhead of model quantization across benchmarks and the BER of the extracted watermark from each quantized model. Note that for all benchmarks, the reported BER is well below the tolerable threshold determined in \textsection \ref{Reliable-Sec}, meaning that watermarks can be verified successfully.

\begin{table}[h!]
  \caption{Robustness of GradSigns against model quantization.\newline}
  \label{quant-table}
  \centering
  \resizebox{\columnwidth}{!}{
    \begin{tabular}{|c|c|c|c|}
    \hline
    \multicolumn{1}{|c|}{} & CIFAR10 & YTF & SVHN \\
    \hline
    \rule{0pt}{10pt}
    BER (\# bit error/total) & 2/64 & 1/64 & 0/64 \\
    \hline
    \rule{0pt}{10pt}
    Performance difference & -1.42\% & -1.73\% & -0.6\% \\
    \hline
    \end{tabular}
    }
\end{table}

\textbf{Adversarial Fine-tuning.} The adversary might perform additional fine-tuning with non-standard objectives to more directly attack the gradient. To this end, we considered fine-tuning benchmark models with adversarial examples which is normally performed for improving the robustness of DNNs against test time adversarial attacks. In this experiment, a set of adversarial examples are generated following the FGSM method \cite{goodfellow2014explaining} and benchmark models are fine-tuned for another 5 epochs with a mixture of generated adversarial examples (with their correct label) and original training samples. As reported in Table \ref{adv-table}, our evaluations showed that watermarks embedded by GradSign remain verifiable after adversarial fine-tuning. Adversarial training resulted in the worst case 5 mismatched bits (i.e. BER of less than 8\%) for the extracted watermark across all benchmarks which is well below the tolerable threshold determined in \textsection \ref{Reliable-Sec}. Table \ref{adv-table} also reports the effect of adversarial fine-tuning on performance of each benchmark. 
\begin{table}[h!]
  \caption{Robustness of GradSigns against adversarial fine-tuning.\newline}
  \label{adv-table}
  \centering
  \resizebox{\columnwidth}{!}{
    \begin{tabular}{|c|c|c|c|}
    \hline
    \multicolumn{1}{|c|}{} & CIFAR10 & YTF & SVHN \\
    \hline
    \rule{0pt}{10pt}
    BER (\# bit error/total) & 5/64 & 0/64 & 0/64 \\
    \hline
    \rule{0pt}{10pt}
    Performance difference & -2.58\% & +0.23\% & -0.22\% \\
    \hline
    \end{tabular}
    }
\end{table}

\textbf{Input Noise Injection} is an example of input tampering attacks in which the adversary places a random noise on queries to the stolen model to corrupt the gradients approximated through zeroth-order methods, and ultimately prevent watermark verification. For this attack, we consider superimposing a Gaussian noise with zero mean and standard deviation varying from 0.001 to 0.1 with inputs to benchmark models. Note that in all our experiments pixel values of images are scaled to $[0.0,1.0]$.

\begin{figure*}[t!]
\centering
\subfigure[64-bits YTF ]{
\includegraphics[width=4cm,height=4cm]{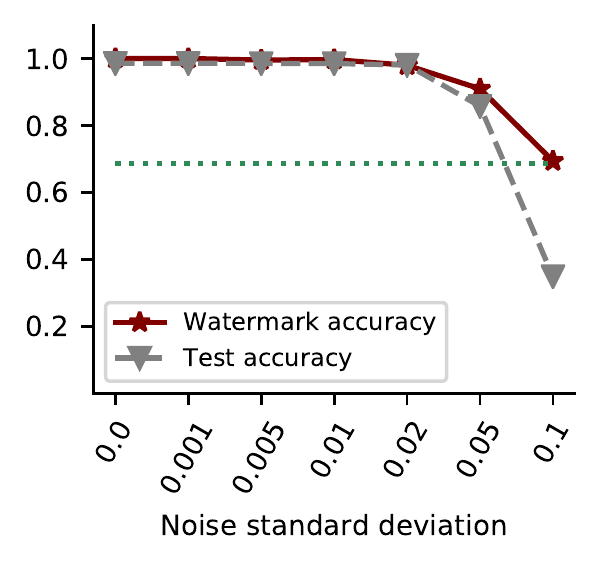}}
\quad
\subfigure[64-bits SVHN ]{
\includegraphics[width=4cm,height=4cm]{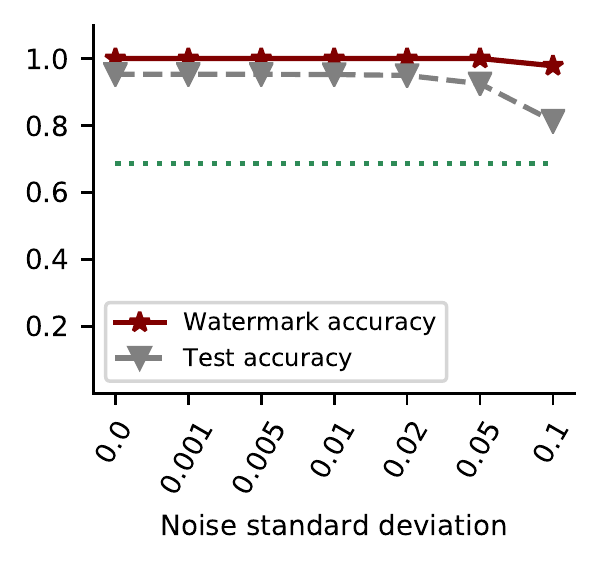}}
\quad
\subfigure[64-bits CIFAR10 ]{
  \includegraphics[width=4cm,height=4cm]{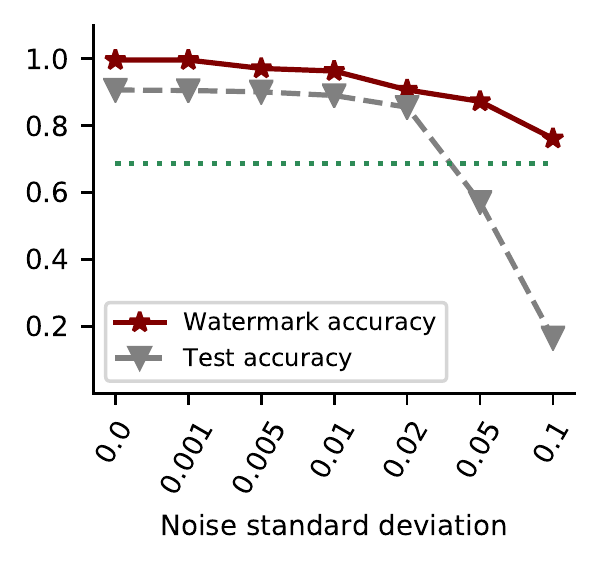}} 
\caption{Resiliency of GradSigns framework against input noise injection attacks. Superimposed noise is drawn from Gaussian distribution $\mu(0.0,\sigma)$ with $\sigma$ varying along the horizontal axis. The green dotted line is the tolerated mismatch threshold.}
\label{INI}
\end{figure*}

Figure \ref{INI} demonstrates the results of our experiment. As shown, watermarks embedded by GradSigns remain existent and verifiable in presence of input noise injection attack. For all cases where the superimposed noise is not too large to heavily degrade the model's performance, watermarks embedded by GradSigns can be successfully verified.

\textbf{Score Rounding} is an example of output tampering attacks in which the adversary aims to break the zeroth-order gradient approximation method in GradSigns' verification phase. As mentioned in \textsection \ref{wm-extraction-section}, estimating the gradient of a carrier node $c$ on input sample $x$ through a zeroth-order method involves querying the model with inputs $x$ and $x+he_{c}$, calculating cross-entropy loss based on the reported prediction probabilities for each query, and finally dividing the observed difference in cross-entropy values by the estimation step length $h$. The adversary, being aware of our framework, could try to mask the difference observed on the output of the model for these two queries by rounding the reported prediction probabilities, falsify the calculated cross-entropy values, and eventually corrupt the estimated gradients.

\begin{figure*}[t!]
\centering
\subfigure[64-bits YTF ]{
\includegraphics[width=4cm,height=4cm]{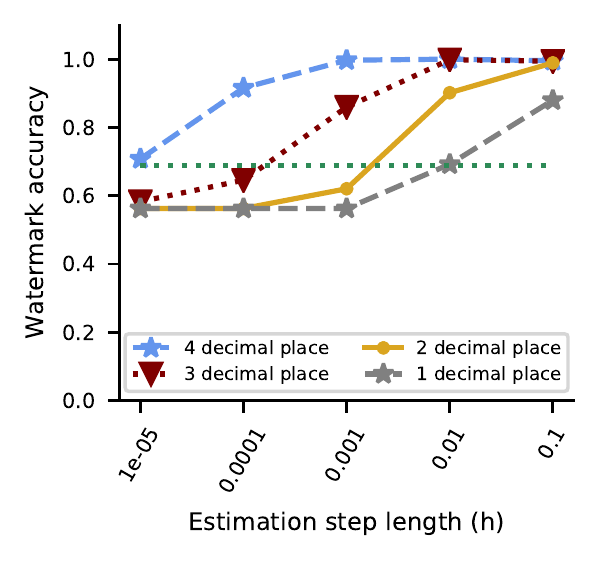}}
\quad
\subfigure[64-bits SVHN ]{
\includegraphics[width=4cm,height=4cm]{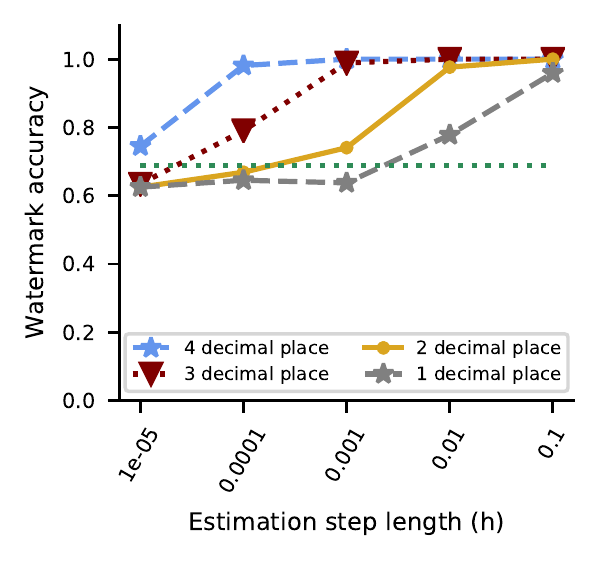}}
\quad
\subfigure[64-bits CIFAR10 ]{
\includegraphics[width=4cm,height=4cm]{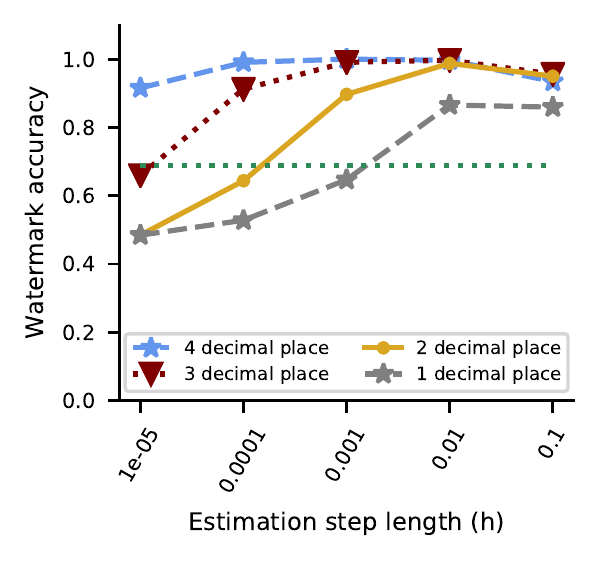}}
\caption{Resiliency of GradSigns framework against score rounding attack. The green dotted line is the tolerated mismatch threshold.}
\label{SR}
\end{figure*}

Figure \ref{SR} demonstrates experimental results for score rounding attack on CIFAR10, SVHN and YTF benchmark models. As shown, in occasions that rounding the reported prediction probabilities can obfuscate the correct value of input gradients, and result in unsuccessful watermark extraction, increasing estimation step length $h$ to values larger than $0.0001$ (which was suggested in \textsection \ref{wm-extraction-section} for watermark extraction in non-adversarial setting) helps GradSigns extract watermark successfully even for cases where the adversary reports model's prediction probabilities with only one decimal place precision (gray lines in Figure \ref{SR}). Using larger estimation step lengths results in larger differences between reported prediction probabilities for inputs $x$ and $x+he_{c}$, which would still remain noticeable over the reported rounded probabilities. The horizontal axis of diagrams in Figure \ref{SR} denotes the value of estimation step length $h$ for the gradient approximation method in GradSigns' verification phase.

\textbf{Score Perturbation} is a more aggressive instance of output tampering attacks in which the adversary deliberately modifies reported prediction probabilities to incur error on the approximated gradients, and prevent watermark verification. While we have considered this category of attacks in our analysis, we believe this category of gradient tampering schemes are less practical for the following reasons: (a) The anti-watermark system can not tell apart queries from the model vendor and regular clients, therefore, regular users will be impacted by fabricated outputs unintentionally, (b) for models deployed in a critical application such as defense, avionics, disease diagnosis in healthcare, etc. exact values of prediction scores are required as direct inputs for policy-based decision-making systems.

We evaluate the performance of GradSigns against score perturbation attacks, in which the adversary introduces random noise on reported prediction probabilities to corrupt computed gradients, and obstruct watermark verification. Our empirical evaluation shows (refer to Appendix \textsection \ref{SPA-app}) that watermarks embedded by GradSigns remain verifiable in presence of this category of attacks.   

\subsection{Credibility}
\label{credibility-Sec}
Credibility requires that finding a ghost (accidental) watermark or forging a secondary watermark into a watermarked model should be impossible meaning that a watermark exists in the model, \emph{if and only if} it has been deliberately embedded by the model vendor. 

If the adversary is able to find a ghost watermark in the stolen model, they can falsely claim that the model belongs to them. To protect GradSigns against \textit{ghosting attacks}, we require the owner to bind their identity to the embedded watermark information. Similar to \cite{Guo}, we suggest that the model owner, instead of using an RNG, generate the watermark $b$ and embedding key $K$ by hashing a meaningful message containing their identity so that even if the adversary is able to find a tuple of $(\hat{b}$\;,\;$\hat{K})$ for which the Equation \ref{embedding-success-eq} holds true, the probability that $\hat{b}$ and $\hat{K}$ are also hash of a meaningful message containing adversary's identity would be close to zero.

Assuming that adversary is aware of our framework, they might try to forge their own watermark, a \textit{counterfeit watermark}, into the vendor's model, and claim its ownership. To perform the \textit{forging attack}, they have to follow the steps 1-4 discussed in \textsection \ref{wm-embedding-section}, and then re-train the stolen model using their training dataset to minimize GradSigns' cost function (Equation \ref{GS-loss}). In this scenario, we seek to answer the following questions, (1) Is it possible to embed a watermark into a pre-trained model? if the answer to the first question is positive, (2) how big of a dataset does an adversary require to forge their own watermark into the model? 

Table \ref{forg-wm} shows the results of a forging attack in which the adversary tries to embed a counterfeit 32-bit signature into the vendor's 32-bit marked SVHN benchmark. We consider forging a counterfeit watermark using sub-datasets of varying sizes to answer the second question. To embed the watermark, the benchmark model is re-trained for another 80 epochs, the same number of epochs required to train the model (from scratch) and embed the original watermark. For each sub-dataset, we conduct the forging attack 3 times with varying $\lambda$s with values smaller, equal, and greater than what was used for embedding the original watermark, and report the best BESR among these trials on Table \ref{forg-wm}. As shown, the adversary is not able to successfully forge their own watermark unless they have access to a dataset with comparable size to the vendor's original dataset, which is considered out-of-scope in our threat model. As we mentioned before, an adversary with access to such a large dataset ($4096\times10 \approx 40k$ samples) has no incentive to hijack the model in the first place as they are capable of training the model from scratch. Even if we assume that the adversary has access to such a dataset, the performance of the model after the forging attempt will be below the stolen model, which points out to inefficiency of forging attacks.

\begin{table}
  \caption{BESR of counterfeit watermark and performance overhead of forging attack. \newline}
  \label{forg-wm}
  \centering
  \resizebox{\columnwidth}{!}{
    \begin{tabular}{|c|c|c|c|c|c|}
    \hline
    \multirow{2}{*}{} & \multicolumn{5}{c|}{Available samples per classification label}\\
    \cline{2-6}
    & 256 & 512 & 1024 & 2048 & \textbf{4096} \\
    \hline
    BESR & 0.57 & 0.60 & 0.62 & 0.68 & \textbf{1.0} \\
    \hline
    Performance overhead & -1.57\% & -1.15\% & -4.43\% & -3.37\% & \textbf{-1.09\%} \\
    \hline
    \end{tabular}
    }
\end{table}

\subsection{Efficiency}
\label{Efficinecy-Sec}
The efficiency metric requires the costs involved in extracting and verifying the watermark to be affordable. This requirement gains importance in scenarios that the stolen model is monetized on an MLaaS \cite{ribeiro2015mlaas} platform where customers are charged per-query basis. For the cases that the stolen model is either deployed on MLaaS platforms with run-time based pricing policy, or is commercialized as an encrypted software, the efficiency metric is not as important since the model vendor is able to make unlimited queries upon purchasing the license. In our method, as mentioned in \textsection \ref{wm-extraction-section}, extracting a watermark involves estimating the expected gradient of the cost function with respect to the carrier nodes, which requires $s \times(|C|+1)$ queries to the model. Here, $|C|$ is the size of the watermark carrier set, and $s$ is the sample size that determines the number of watermark key images over which the expected gradients are estimated. The standard error of the approximated expected gradients, is directly affected by the sample size, in that, the larger the sample size, the smaller the error would be \cite{probab}.

Note that there is a trade-off between efficiency and accuracy of watermark extraction. On one hand, to successfully extract the watermark, the model owner needs to compute the expected input gradients with high accuracy which requires a large set of watermark key images, as mentioned before. On the other hand, a large sample size translates to higher verification costs for the vendor. Therefore, to balance the efficiency and accuracy of watermark extraction, the model vendor needs to determine the minimum number of watermark key images (minimum sample size) that guarantees a successful watermark extraction. Column 4 of Table \ref{efficinecy-table} reports the minimum sample size required for extraction of the watermark from CIFAR-10, SVHN and YTF benchmark models. These numbers have been experimentally proven to be large enough for a successful watermark extraction over 20 trials. Watermark size, watermark carrier set size and number of extraction queries per watermark bit for different benchmarks are listed on Table \ref{efficinecy-table}. Number of verification queries can be reduced by using random-direction based gradient estimation \cite{liu2018signsgd,tu2018autozoom} instead of the coordinate-wise estimation method as in Equation \ref{zeroth-order-grad-est-Equation}.

\subsection{Capacity}
Capacity requires the watermarking method to be able to embed large amount of information into the model. As shown in Table \ref{benchmarks-test-acc-table}, unlike contemporary black-box methods which can only embed one bit of information, GradSigns is able to embed watermarks with various size of 16, 32 and 64 bits into all CIFAR-10, SVHN and YTF model benchmarks, which shows our method fulfills this requirement.

\begin{table}
  \caption{Efficiency of extracting watermark with GradSigns. \newline}
  \label{efficinecy-table}
  \centering
  \scalebox{0.60}{
    \begin{tabular}{|c|c|c|c|c|}
    \hline
    Dataset & WM Size & Carrier set size   & Min. sample size &  extraction queries per WM bit   \\
     \hline
    \multirow{3}{*}{CIFAR-10}  &  16  & 128  & \multirow{4}{*}{50} & \multirow{4}{*}{$<500$}  \\
    \cline{2-3}  
     & 32 & 256 & & \\
     \cline{2-3}  
     & 64 & 384 & &  \\
    \hline
    \multirow{3}{*}{SVHN}  & 16 & 256 & \multirow{4}{*}{50} & \multirow{4}{*}{$<500$}   \\
    \cline{2-3}  
     & 32 & 256 & &   \\
     \cline{2-3}  
     & 64 & 512 & &   \\
    \hline
    \multirow{3}{*}{YTF}  & 16  & 64 & \multirow{4}{*}{20} & \multirow{4}{*}{$<100$}  \\
    \cline{2-3}  
     & 32 & 128 & &  \\
     \cline{2-3}  
     & 64 & 128 & &  \\
    \hline
    \end{tabular}
    }
\end{table}

\section{Conclusion}
In this paper, we present GradSigns, a novel watermarking framework for DNNs. GradSigns embeds the owner's signature by imposing a statistical bias on the expected gradient of the cost function with respect to the model's input. We evaluate GradSigns on DNNs trained with three different image datasets, and experimentally show that (1) our method is extremely robust against general and adaptive counter-watermark attacks (2) it is capable of embedding a large amount of information with negligible impact on the performance of the model, and (3) it is efficient, reliable and credible, thereby providing a strong proof of ownership and a practical tool for supporting responsible AI technology.

\bibliography{paper}

\begin{thebibliography}{31}
\providecommand{\natexlab}[1]{#1}
\providecommand{\url}[1]{\texttt{#1}}
\expandafter\ifx\csname urlstyle\endcsname\relax
  \providecommand{\doi}[1]{doi: #1}\else
  \providecommand{\doi}{doi: \begingroup \urlstyle{rm}\Url}\fi

\bibitem[Adi et~al.(2018)Adi, Baum, Cisse, Pinkas, and Keshet]{adi2018turning}
Adi, Y., Baum, C., Cisse, M., Pinkas, B., and Keshet, J.
\newblock Turning your weakness into a strength: Watermarking deep neural
  networks by backdooring.
\newblock In \emph{27th USENIX Security Symposium (USENIX Security 18)}, pp.\
  1615--1631, 2018.

\bibitem[Allen-Zhu et~al.(2019)Allen-Zhu, Li, and Liang]{allen2019learning}
Allen-Zhu, Z., Li, Y., and Liang, Y.
\newblock Learning and generalization in overparameterized neural networks,
  going beyond two layers.
\newblock In \emph{Advances in neural information processing systems}, pp.\
  6155--6166, 2019.

\bibitem[Borra et~al.(2018)Borra, Thanki, and Dey]{borra2018digital}
Borra, S., Thanki, R., and Dey, N.
\newblock \emph{Digital Image Watermarking: Theoretical and Computational
  Advances}.
\newblock Intelligent Signal Processing and Data Analysis. CRC Press, 2018.

\bibitem[Dekel et~al.(2017)Dekel, Rubinstein, Liu, and Freeman]{visionWM}
Dekel, T., Rubinstein, M., Liu, C., and Freeman, W.~T.
\newblock On the effectiveness of visible watermarks.
\newblock In \emph{Proceedings of the IEEE Conference on Computer Vision and
  Pattern Recognition}, pp.\  2146--2154, 2017.

\bibitem[Du \& Lee(2018)Du and Lee]{du2018power}
Du, S. and Lee, J.
\newblock On the power of over-parametrization in neural networks with
  quadratic activation.
\newblock In \emph{Proceedings of the 35th International Conference on Machine
  Learning}, volume~80 of \emph{Proceedings of Machine Learning Research}, pp.\
   1329--1338, 10--15 Jul 2018.

\bibitem[Ghadimi \& Lan(2013)Ghadimi and Lan]{ghadimi2013stochastic}
Ghadimi, S. and Lan, G.
\newblock Stochastic first-and zeroth-order methods for nonconvex stochastic
  programming.
\newblock \emph{SIAM Journal on Optimization}, 23\penalty0 (4):\penalty0
  2341--2368, 2013.

\bibitem[Goodfellow et~al.(2014)Goodfellow, Shlens, and
  Szegedy]{goodfellow2014explaining}
Goodfellow, I.~J., Shlens, J., and Szegedy, C.
\newblock Explaining and harnessing adversarial examples.
\newblock \emph{arXiv preprint arXiv:1412.6572}, 2014.

\bibitem[Guo \& Potkonjak(2018)Guo and Potkonjak]{Guo}
Guo, J. and Potkonjak, M.
\newblock Watermarking deep neural networks for embedded systems.
\newblock In \emph{Proceedings of the International Conference on
  Computer-Aided Design}, ICCAD '18, pp.\  133:1--133:8. ACM, 2018.

\bibitem[Han et~al.(2015)Han, Pool, Tran, and Dally]{han2015learning}
Han, S., Pool, J., Tran, J., and Dally, W.
\newblock Learning both weights and connections for efficient neural network.
\newblock In \emph{Advances in neural information processing systems}, pp.\
  1135--1143, 2015.

\bibitem[He et~al.(2016)He, Zhang, Ren, and Sun]{he2016deep}
He, K., Zhang, X., Ren, S., and Sun, J.
\newblock Deep residual learning for image recognition.
\newblock In \emph{Proceedings of the IEEE conference on computer vision and
  pattern recognition}, pp.\  770--778, 2016.

\bibitem[Krizhevsky et~al.()Krizhevsky, Nair, and Hinton]{cifar10}
Krizhevsky, A., Nair, V., and Hinton, G.
\newblock Cifar-10 (canadian institute for advanced research).
\newblock URL \url{http://www.cs.toronto.edu/~kriz/cifar.html}.

\bibitem[Le~Merrer et~al.(2019)Le~Merrer, Perez, and Tr{\'e}dan]{merrer}
Le~Merrer, E., Perez, P., and Tr{\'e}dan, G.
\newblock Adversarial frontier stitching for remote neural network
  watermarking.
\newblock \emph{Neural Computing and Applications}, pp.\  1--12, 2019.

\bibitem[LeCun \& Cortes(2010)LeCun and
  Cortes]{lecun-mnisthandwrittendigit-2010}
LeCun, Y. and Cortes, C.
\newblock {MNIST} handwritten digit database.
\newblock 2010.
\newblock URL \url{http://yann.lecun.com/exdb/mnist/}.

\bibitem[Liu et~al.(2018)Liu, Kailkhura, Chen, Ting, Chang, and
  Amini]{liu2018zeroth}
Liu, S., Kailkhura, B., Chen, P.-Y., Ting, P., Chang, S., and Amini, L.
\newblock Zeroth-order stochastic variance reduction for nonconvex
  optimization.
\newblock In \emph{Advances in Neural Information Processing Systems}, pp.\
  3727--3737, 2018.

\bibitem[Liu et~al.(2019)Liu, Chen, Chen, and Hong]{liu2018signsgd}
Liu, S., Chen, P.-Y., Chen, X., and Hong, M.
\newblock sign{SGD} via zeroth-order oracle.
\newblock In \emph{International Conference on Learning Representations}, 2019.

\bibitem[Liu et~al.(2020)Liu, Chen, Kailkhura, Zhang, Hero, and
  Varshney]{liu2020primer}
Liu, S., Chen, P.-Y., Kailkhura, B., Zhang, G., Hero, A., and Varshney, P.~K.
\newblock A primer on zeroth-order optimization in signal processing and
  machine learning.
\newblock \emph{IEEE Signal Processing Magazine}, 2020.

\bibitem[Namba \& Sakuma(2019)Namba and Sakuma]{namba2019robust}
Namba, R. and Sakuma, J.
\newblock Robust watermarking of neural network with exponential weighting.
\newblock In \emph{Proceedings of the 2019 ACM Asia Conference on Computer and
  Communications Security}, pp.\  228--240, 2019.

\bibitem[Netzer et~al.(2011)Netzer, Wang, Coates, Bissacco, Wu, and Ng]{SVHN}
Netzer, Y., Wang, T., Coates, A., Bissacco, A., Wu, B., and Ng, A.~Y.
\newblock Reading digits in natural images with unsupervised feature learning.
\newblock \emph{NIPS Workshop on Deep Learning and Unsupervised Feature
  Learning}, 2011.

\bibitem[Qu \& Potkonjak(1998)Qu and Potkonjak]{Qu98}
Qu, G. and Potkonjak, M.
\newblock Analysis of watermarking techniques for graph coloring problem.
\newblock \emph{Proceedings of the 1998 IEEE/ACM international conference on
  Computer-aided design}, pp.\  190--193, 1998.

\bibitem[Qu \& Potkonjak(2003)Qu and Potkonjak]{Qu_book03}
Qu, G. and Potkonjak, M.
\newblock Intellectual property protection in vlsi designs: Theory and
  practice.
\newblock \emph{Kluwer Academic Publishers}, 2003.

\bibitem[Ribeiro et~al.(2015)Ribeiro, Grolinger, and Capretz]{ribeiro2015mlaas}
Ribeiro, M., Grolinger, K., and Capretz, M.~A.
\newblock Mlaas: Machine learning as a service.
\newblock In \emph{IEEE 14th International Conference on Machine Learning and
  Applications (ICMLA)}, pp.\  896--902. IEEE, 2015.

\bibitem[Rouhani et~al.(2018)Rouhani, Chen, and
  Koushanfar]{rouhani2018deepsigns}
Rouhani, B.~D., Chen, H., and Koushanfar, F.
\newblock Deepsigns: A generic watermarking framework for ip protection of deep
  learning models.
\newblock \emph{arXiv preprint arXiv:1804.00750}, 2018.

\bibitem[Saini \& Shrivastava(2014)Saini and Shrivastava]{saini2014survey}
Saini, L.~K. and Shrivastava, V.
\newblock A survey of digital watermarking techniques and its applications.
\newblock \emph{arXiv preprint arXiv:1407.4735}, 2014.

\bibitem[Sun et~al.(2014)Sun, Wang, and Tang]{sun2014deep}
Sun, Y., Wang, X., and Tang, X.
\newblock Deep learning face representation from predicting 10,000 classes.
\newblock In \emph{Proceedings of the IEEE conference on computer vision and
  pattern recognition}, pp.\  1891--1898, 2014.

\bibitem[Tamhane \& Dunlop(2000)Tamhane and Dunlop]{probab}
Tamhane, A. and Dunlop, D.
\newblock \emph{Statistics and Data Analysis: From Elementary to Intermediate}.
\newblock Prentice Hall, 2000.

\bibitem[Tu et~al.(2019)Tu, Ting, Chen, Liu, Zhang, Yi, Hsieh, and
  Cheng]{tu2018autozoom}
Tu, C.-C., Ting, P., Chen, P.-Y., Liu, S., Zhang, H., Yi, J., Hsieh, C.-J., and
  Cheng, S.-M.
\newblock Autozoom: Autoencoder-based zeroth order optimization method for
  attacking black-box neural networks.
\newblock In \emph{Proceedings of the AAAI Conference on Artificial
  Intelligence}, volume~33, pp.\  742--749, 2019.

\bibitem[Uchida et~al.(2017)Uchida, Nagai, Sakazawa, and Satoh]{Uchida_2017}
Uchida, Y., Nagai, Y., Sakazawa, S., and Satoh, S.
\newblock Embedding watermarks into deep neural networks.
\newblock \emph{Proceedings of the 2017 ACM on International Conference on
  Multimedia Retrieval - ICMR ’17}, 2017.

\bibitem[Wolf et~al.(2011)Wolf, Hassner, and Maoz]{wolf2011face}
Wolf, L., Hassner, T., and Maoz, I.
\newblock \emph{Face recognition in unconstrained videos with matched
  background similarity}.
\newblock IEEE, 2011.

\bibitem[Wong et~al.(2004)Wong, Qu, and Potkonjak]{Wong04}
Wong, J.~L., Qu, G., and Potkonjak, M.
\newblock Optimization-intensive watermarking techniques for decision problems.
\newblock \emph{IEEE Transactions on Computer-Aided Design of Integrated
  Circuits and Systems}, 23\penalty0 (1):\penalty0 119--127, 2004.

\bibitem[Zhang et~al.(2018)Zhang, Gu, Jang, Wu, Stoecklin, Huang, and
  Molloy]{Zhang}
Zhang, J., Gu, Z., Jang, J., Wu, H., Stoecklin, M.~P., Huang, H., and Molloy,
  I.
\newblock Protecting intellectual property of deep neural networks with
  watermarking.
\newblock In \emph{Proceedings of the 2018 on Asia Conference on Computer and
  Communications Security}, pp.\  159--172, 2018.

\bibitem[Zhu et~al.(2005)Zhu, Thomborson, and Wang]{softwarewmsurvey}
Zhu, W., Thomborson, C., and Wang, F.-Y.
\newblock A survey of software watermarking.
\newblock In Kantor, P., Muresan, G., Roberts, F., Zeng, D.~D., Wang, F.-Y.,
  Chen, H., and Merkle, R.~C. (eds.), \emph{Intelligence and Security
  Informatics}, pp.\  454--458. Springer Berlin Heidelberg, 2005.

\end{thebibliography}
\bibliographystyle{mlsys2021}

\appendix
\clearpage
\section{More on Query invalidation and modification}
\label{qim-app}
Query invalidation and modification \cite{namba2019robust} is a counter watermark attack in which the adversary utilizes an autoencoder to identify and modify model queries that are executed to extract the watermark. Namba et al. \cite{namba2019robust} argue that if an input $x$ is drawn from the same distribution as the training data, the reconstruction loss of input $x$ introduced by any auto encoder $AE$ trained on the same dataset should be relatively smaller compared to an input sample coming from a different distribution (e.g. a sample created by superimposing images with noise or a context). In this approach, besides the reconstruction loss introduced by the autoencoder, the Jensen-Shanon divergence between predicted class probabilities of the model for input $x$ and the reconstructed input $AE(x)$ is used as a measure to differentiate ordinary input samples from watermark key set. This distance is larger for key samples compared to ordinary inputs, as most of existing watermarking methods label watermark key samples differently from their original classes. In this approach, if input sample $x$ is identified as suspicious, i.e. possibly belonging to the vendor's watermark key set, the model will be queried using $AE(x)$ instead.
While this counter watermark attack has proven effective \cite{namba2019robust} against watermarking techniques proposed by \citet{Zhang} and \citet{rouhani2018deepsigns}, it doesn't apply to GradSigns. The watermark key set for GradSigns is constructed from samples of training data without any modification or relabeling, which renders this attack futile against our method.

\section{Score Perturbation Attacks}
\label{SPA-app}
In this class of adaptive attacks, the adversary has two goals: (1) maximally perturbing the probability of top-1 predicted label to corrupt the approximated gradients, (2) minimizing impacts of output perturbations on the model's functionality to preserve its competitive performance and reduce possible harm to non-vendor users. Achieving the first goal equals maximizing the L-infinity ($L\infty$) norm of output modifications. To accomplish the second goal, an adversary should (a) modify as few prediction scores as possible to preserve the fidelity of the reported outputs, and (b) maintain the true order of top-$M$ predicted labels in the fabricated outputs. The latter property is referred to as \textit{$M$-true label ordering} in the rest of the paper. In our experiments, $M$ is set to 3.

In the Score Perturbation (SP) attack, the prediction probability of the top-1 label is superimposed with random noise to corrupt cross-entropy values computed for approximating the gradients of carrier nodes. To keep the sum of reported prediction probabilities equal to 1.0, the reported score of other label(s) should also be modified. The adversary, to minimize impacts of perturbations on the fidelity of reported outputs, only modifies (increases) the prediction probability of one other output label, preferably the label with the least probability.

Algorithm \ref{non-deterministic-alg} summarizes the logistics of SP attack. In lines 2 and 3 of this algorithm, input sample $X$ is fed to the model, and the maximum possible value $\tau_{sp}$ which can be reduced from the probability of top-1 label is computed. In our notation, $P_{top_{j}}$ indicates the $j_{th}$ highest probability in $P$. Value of $\tau_{sp}$ is upper bounded by ($P_{top_{M}}$-$P_{top_{|P|}}$) and ($P_{top_{1}}$-$P_{top_{2}}$) to assure that the $M$-true label ordering property is held true on the fabricated outputs. In line 4, magnitude (L-infinity norm) of perturbations is randomly decided from a uniform distribution on $(0,\tau_{sp}]$ and in lines 5-6, prediction scores of the most and least likely labels ($P_{top_{1}} $ and $P_{top_{|P|}}$) are modified correspondingly. 

\begin{figure*}[t]
\centering
\subfigure[CIFAR10 benchmarks]{
  \includegraphics[width=4cm,height=4cm]{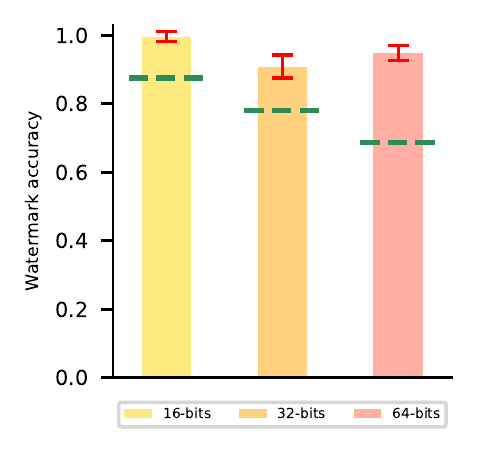}} 
\subfigure[SVHN benchmarks]{
  \includegraphics[width=4cm,height=4cm]{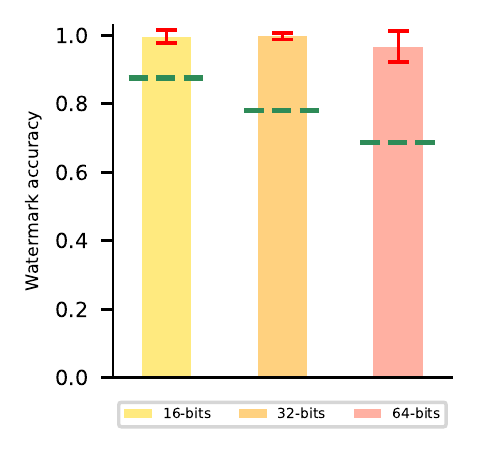}}
\subfigure[YTF benchmarks]{
  \includegraphics[width=4cm,height=4cm]{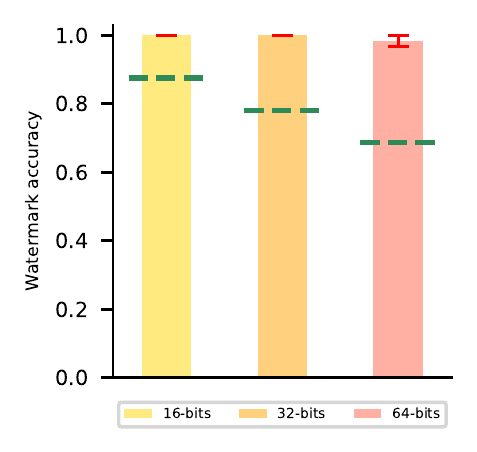}}
\caption{Accuracy of extracted watermark in presence of score perturbation attack. The green dashed line is the tolerated mismatch threshold for each benchmark.}
\label{NSP}
\end{figure*}

Figure \ref{NSP} demonstrates results of score perturbation attack for all three datasets. As shown, this attack fails to prevent the model vendor from verifying the embedded watermark as the accuracy of the extracted watermarks are always above the accepted threshold derived in \textsection \ref{Reliable-Sec}.
In this experiment, for each benchmark, 50 verification attempts in presence of SP attack are made, and the mean and standard deviation of the accuracy of extracted watermark over these trials are reported on diagrams in Figure \ref{NSP}. The estimation step length is set to $0.1$ throughout this experiment for all benchmarks, and $\epsilon$ is a small constant set to $0.00001$.

\begin{algorithm}[t]
\caption{Score perturbation attack}
\label{non-deterministic-alg}
\begin{algorithmic}[1]
\STATE {\bfseries Input:} input sample $X$, model $\Bar{H}$, true label ordering $M$
\STATE P = $\Bar{H}(X)$
\STATE $\tau_{sp} = \min\{(P_{top_{1}}-P_{top_{2}})-\epsilon,(P_{top_{M}}-P_{top_{|P|}})-\epsilon\}$
\STATE $\phi  \leftarrow$ $uniform(0,\tau_{sp})$
\STATE $P_{top_{|P|}}\,=\,P_{top_{|P|}}+\phi$
\STATE $P_{top_{1}}\,=\,P_{top_{1}}-\phi$
\STATE {\bfseries Output:} $P$
\end{algorithmic}
\end{algorithm}

\begin{table*}[]
  \caption{Implementation details of Zhang et al. \cite{Zhang} method.}
  \label{zhangdet}
  \centering
  \resizebox{\textwidth}{!}{
  \begin{tabular}{|c|c|c|c|c|}
     \hline
    Dataset & Watermark type & Key samples     & \# of key samples     & Label for key samples \\
     \hline
    \multirow{3}{*}{CIFAR-10} & Content & CIFAR-10 airplanes superimposed with string "TEST" in gray & 5000 & cars \\
    \cline{2-5}
     & Unrelated & MNIST\cite{lecun-mnisthandwrittendigit-2010} "1"  & 6000 & cars \\
    \cline{2-5}
     & Noise & CIFAR-10 random superimposed with Gaussian noise $\mu(0,\,0.005)$ & 3000 & cars \\
    
      \hline
    \multirow{3}{*}{SVHN} & Content & SVHN "4" superimposed with string "TEST" in gray & 5000 & SVHN "1" \\
    \cline{2-5}
     & Unrelated & CIFAR-10 cars & 5000 & SVHN "1" \\
    \cline{2-5}
     & Noise & SVHN random superimposed with Gaussian noise $\mu(0,\,0.005)$ & 1000 & SVHN "1" \\
    
        \hline
    \multirow{3}{*}{YTF} & Content & YTF individual \#4 superimposed with string "TEST" in gray & 81 & YTF individual \#1 \\
    \cline{2-5}
     & Unrelated & CIFAR-10 deer & 5000 & YTF individual \#1 \\
    \cline{2-5}
     & Noise & YTF random superimposed with Gaussian noise $\mu(0,\,0.005)$ & 1000 & YTF individual \#1 \\
    \hline
  \end{tabular}
  }
\end{table*}

\section{Datasets and Experimental Setup}
\label{dataset_app}
\textbf{CIFAR-10} \cite{cifar10} is an object classification dataset containing 50,000 training and 10,000 testing samples belonging to 10 different classes. The DNN model used for this dataset is ResNet20 \cite{he2016deep}. 
\textbf{YouTube Face} (YTF) \cite{wolf2011face} is a face recognition dataset containing images of 1,595 individuals captured from videos on YouTube. We retrieve 120,000 images of 1200 individuals (100 images per individual), and use 75\% of the images for training and the remaining as the test set. In our experiments, each image has been resized to 40x40x3 dimension, and the DNN used for this dataset is the state-of-the-art DeepID \cite{sun2014deep}. \textbf{SVHN} \cite{SVHN} is a dataset of more than 100k images of digits cropped out of images of houses and street numbers. The dimension of each image in this dataset is 32x32x3. The architecture of the convolutional neural network considered for this dataset is reported in Table \ref{SVHN-ARCH}. 
For methods AFS-\cite{merrer}, DS-\cite{rouhani2018deepsigns}, and EXP-\cite{namba2019robust}, we have embedded 30 key samples by fine-tuning the host model as instructed in the original work. Note that all the parameters regarding the watermark embedding procedure were adopted from the original papers. The details of watermark embedding using Zhang-\cite{Zhang} are reported in Table \ref{zhangdet}. 

\begin{table}[h!]
  \caption{Architecture of SVHN Benchmark.}
  \label{SVHN-ARCH}
  \centering
  \resizebox{0.7\columnwidth}{!}{
  \begin{tabular}{cc}
    \hline
    \multicolumn{2}{c}{SVHN}      \\
    \cmidrule(r){1-2}  
    Layer Type     & Filter/Unit \\
    \midrule
    Convolution + ReLU & $3 \times 3 \times 32$    \\
    Convolution + ReLU & $3 \times 3 \times 32$    \\
    Convolution + ReLU & $3 \times 3 \times 64$    \\
    Convolution + ReLU & $3 \times 3 \times 64$    \\
    Convolution + ReLU & $3 \times 3 \times 128$    \\
    Convolution + ReLU & $3 \times 3 \times 128$    \\
    Fully Connected + ReLU & 512 \\
    Softmax & 10 \\
    \bottomrule
  \end{tabular}}
\end{table}

\end{document}